\documentclass[letterpaper, 10 pt, conference]{ieeeconf}  

\IEEEoverridecommandlockouts                              

\usepackage{hyperref}

\usepackage[noend]{algpseudocode}
\usepackage{diagbox} 
\usepackage{amsmath}
\usepackage{cleveref}
\usepackage{booktabs}
\usepackage{balance}
\usepackage{multirow}
\usepackage{gensymb} 

\usepackage{graphicx}
\usepackage{amssymb}
\usepackage[caption=false,font=scriptsize,labelfont=sf,textfont=sf]{subfig}

\title{\LARGE \bf
Unlocking Thermal Aerial Imaging: Synthetic Enhancement of UAV Datasets
}

\author{Antonella Barisic Kulas$^{1}$, Andreja Jurasovic and Stjepan Bogdan$^{1}$
\thanks{$^{1}$Authors are with the University of Zagreb Faculty of Electrical Engineering  and Computing, LARICS Laboratory for Robotics and Intelligent Control Systems, Unska 3, Zagreb 10000, Croatia; {\tt\small (authors)@fer.unizg.hr}}}

\begin{document}

\maketitle
\thispagestyle{empty}
\pagestyle{empty}

\begin{abstract}

Thermal imaging from unmanned aerial vehicles (UAVs) holds significant potential for applications in search and rescue, wildlife monitoring, and emergency response, especially under low-light or obscured conditions. However, the scarcity of large-scale, diverse thermal aerial datasets limits the advancement of deep learning models in this domain, primarily due to the high cost and logistical challenges of collecting thermal data. In this work, we introduce a novel procedural pipeline for generating synthetic thermal images from an aerial perspective. Our method integrates arbitrary object classes into existing thermal backgrounds by providing control over the position, scale, and orientation of the new objects, while aligning them with the viewpoints of the background. We enhance existing thermal datasets by introducing new object categories, specifically adding a drone class in urban environments to the HIT-UAV dataset and an animal category to the MONET dataset. In evaluating these datasets for object detection task, we showcase strong performance across both new and existing classes, validating the successful expansion into new applications. Through comparative analysis, we show that thermal detectors outperform their visible-light-trained counterparts and highlight the importance of replicating aerial viewing angles. Project page: \url{https://github.com/larics/thermal_aerial_synthetic}.

\end{abstract}


\section{Introduction}
\label{sec:intro}

Flying robots, commonly known as UAVs or drones, have revolutionized remote sensing\cite{Osco2021}. Their potential impact goes far beyond that, promising transformative applications in search and rescue operations \cite{Paulin2024}, wildlife protection \cite{Anbalagan2023, 10226334}, environmental monitoring \cite{Farinha2020, Aucone2023} and much more. The key to these advances is the ability to sense and understand the environment. While visible-light cameras have dominated UAV sensing, capturing only a portion of the electromagnetic spectrum, thermal cameras detect infrared radiation, offering unique advantages that visible-light cameras cannot. They enable vision in low-light or obscured conditions, facilitate temperature-based analysis and open up new possibilities for detecting objects based on thermal signatures. In this paper, we explore thermal aerial imaging to aid new horizons for UAV applications.

Deep learning has emerged as the dominant approach for processing and analyzing aerial imagery, but it comes with a well-known challenge: data dependency. Successful deep learning models require large, high-quality datasets to perform accurately. However, obtaining thermal aerial images is expensive and time-consuming, compounded by the high cost of thermal cameras. 


To address this challenge, we propose an innovative solution: leveraging synthetic data augmentation to enhance existing thermal datasets, thereby enabling deep learning models to be trained for various tasks without the logistical burden of manual data collection. We model objects of interest in the virtual scene, diversify their positions and orientations, scale and viewpoint, and create a synthetic thermal image of those objects, which we combine with real thermal backgrounds from already existing datasets. This approach not only mitigates the scarcity of thermal data but also overcomes the limited variety of classes in current thermal aerial datasets, which predominantly feature only people, vehicles, and occasionally animals. Consequently, thermal aerial applications \cite{6907094, Anbalagan2023,10226334} have been restricted to these subjects. Our method removes this limitation, unlocking endless possibilities for thermal imagery and expanding the scope of UAV applications.

\begin{table*}[htb!]
\centering
\caption{Publicly available thermal aerial datasets of real-world images. 1K=1000, LWIR= Long wave infrared, V=Visible spectrum.}
\renewcommand{\arraystretch}{0.8}
\label{tab:datasets}
\begin{tabular}{llcccccc}
\toprule
\textbf{Dataset Attribute} & & \textbf{BIRDSAI} \cite{birdsai} & \textbf{WiSARD} \cite{Broyles2022} & \textbf{HIT-UAV} \cite{Suo2023} & \textbf{MONET} \cite{Riz2023} & \textbf{WIT-UAS} \cite{wit-uas} \\
\midrule
\# Images           &    & 62K & 15K & 2.9K & 53K & 7K \\ 
\# Bounding Boxes   &    & 154K & ? & 25K & 162K & 6.5K \\ 
Resolution          &    & 640x480 & 640x512 & 640x512 & 800x600 & 640x512, 320x240 \\ 
Metadata            &    & -- & \checkmark & \checkmark & \checkmark & \checkmark \\ 
\midrule
\textbf{Categories} & People & \checkmark & \checkmark & \checkmark & \checkmark & \checkmark \\ 
                    & Vehicles & -- & -- & \checkmark & \checkmark & \checkmark \\ 
                    & Animals  & \checkmark & -- & -- & -- & -- \\ 
\midrule
\textbf{Sensor Modality} & & LWIR & V, LWIR & LWIR & LWIR & LWIR \\ 
\midrule
\textbf{Annotation} & AABB   & \checkmark & \checkmark & \checkmark & \checkmark & \checkmark \\ 
                    & OBB    & -- & -- & \checkmark & -- & -- \\ 
                    & Masks  & -- & -- & -- & -- & -- \\ 
                    & Object Trajectories & \checkmark & -- & -- & \checkmark & -- \\ 
\midrule
\textbf{Scenario}   & Urban  & -- & -- & \checkmark & -- & -- \\ 
                    & Rural  & \checkmark & \checkmark & -- & \checkmark & \checkmark \\ 
\midrule
\textbf{Time}       & Day    & -- & \checkmark & \checkmark & \checkmark & \checkmark \\ 
                    & Night  & \checkmark & \checkmark & \checkmark & \checkmark & -- \\ 
\midrule
\textbf{Camera Angle} & & ? & varied & 30\degree  to 90\degree & -40\degree  to 92\degree  & 30\degree  \& 90\degree  \\ 
\midrule
\textbf{Altitude}   & & 60-120 & 20-120 & 60-130 & 20-130 & 20-100 \\ 
\midrule
\textbf{Year}       &  & 2020 & 2022 & 2023 & 2023 & 2023 \\ 
\bottomrule
\end{tabular}
\end{table*}

The main contributions of this paper are:

\begin{enumerate}
    \item \textbf{A novel procedural pipeline for generating synthetic thermal images from an aerial perspective.} To the best of our knowledge, this is the first pipeline of its kind, which:
    \begin{itemize}
        \item Is capable of adding any type of object into the thermal scene.
        \item Can match the viewing angle of the provided thermal backgrounds.
        \item Generates axis-aligned bounding boxes, oriented bounding boxes, and segmentation masks.
    \end{itemize}
    \item \textbf{Enhancement and publication of extended thermal datasets:} \textbf{a) HIT-UAV-drone:} We introduce a new class, the drone, to the HIT-UAV \cite{Suo2023} dataset. This addition addresses critical applications as drones become integral to emergency response, traffic management, and delivery in urban settings. We double the size of the dataset, demonstrating ability to also extend the quantity of the data.
    \textbf{b) MONET-deer:} We add a new animal category to the MONET \cite{Riz2023} dataset, specifically deer, illustrating the capability to introduce new species and supporting applications in wildlife preservation and monitoring.

    \item We demonstrate the usability of our new datasets for object detection tasks using the YOLOv8 object detector. We also compare the performance of thermal detectors with that of detectors trained on visible light data. Furthermore, we analyze how the introduction of new classes affects the performance of existing classes and evaluate the importance of the aerial perspective for improving detection performance.
\end{enumerate}

\section{Related work}
\label{sec:relatedwork}

\textbf{Thermal Aerial Datasets.} With the growing accessibility of thermal technology, there has been a surge in UAV-collected thermal data, leading to the creation of several public datasets. We present an overview in \Cref{tab:datasets}. One of the first thermal datasets recorded from outdoor UAV is \textbf{HIT-UAV} \cite{Suo2023}. This dataset contains aerial images taken at different altitudes, with a focus on higher altitudes, providing a representative of typical aerial images that are characterized by a wide coverage area but reduced object- level detail. Subsequent datasets follow a similar approach, incorporating high-altitude imagery. The HIT-UAV dataset contains images captured in urban environments such as streets, parks and parking lots. The data is manually annotated with two types of annotations: axis-aligned and oriented bounding boxes, with common urban categories: person, car, bicycle and other vehicle. Another dataset, \textbf{MONET} \cite{Riz2023}, focuses on similar categories, especially people and vehicles, and also provides additional metadata. MONET captures scenes in rural areas and thus provides a different context in contrast to the urban focus of HIT-UAV. \textbf{BIRDSAI} \cite{birdsai} is a dataset of thermal images of humans and animals recorded in South Africa, primarily for research in animal protection. The data was recorded with a fixed-wing UAV during night missions and includes trajectory information, but no information about the flight altitude and camera perspective. Another data set that focuses on the wilderness, \textbf{WiSARD} \cite{Broyles2022}, is dedicated to human detection and provides synchronized visual- thermal image pairs across diverse environmental conditions. The \textbf{WiT-UAS} \cite{wit-uas} dataset, on the other hand, focuses on assets such as vehicles in the vicinity of wildfires, highlighting the critical role of context in distinguishing these objects from surrounding fire and heat. While these datasets contribute significantly to the field, they are largely centered on a limited range of categories, primarily persons, vehicles, and to a lesser extent, animals. There is a clear need to expand thermal aerial datasets to cover a broader range of object categories to improve the development and scope of thermal imaging models.

\textbf{Thermal Synthetic Datasets.} In visual imagery, the lack of real data and hard to obtain or expensive to obtain data can be solved by producing synthetic images, either using virtual world \cite{Barisic2022, Denninger2023} or generative models. The topic of synthetic thermal imagery is rather new, and especially in robotics, and aerial imagery. One of the first works is BIRDSAI, which besides real data contains also synthetic portion. In \cite{birdsai}, the authors use simulation platform AirSim-W \cite{Bondi2018} to generate synthetic aerial TIR videos. Platform has TIR camera model, with resolution $640\times480$. In \cite{Madan2023}, the authors create a pipeline where they use real thermal backgrounds, recreate part of the world in simulation, and simulate people falling into the water. This shows the power of synthetic in cases when real data is hard and dangerous to obtain. Their dataset contains static camera, and single camera view, which is more focused towards video surveillance. We diversify our viewpoints, aiming to cover full aerial viewpoint range, to achieve better performance in aerial applications.



\section{Method}
\label{sec:method}

\begin{figure*}[htb]
  \centering
   \includegraphics[width=1.0\linewidth]{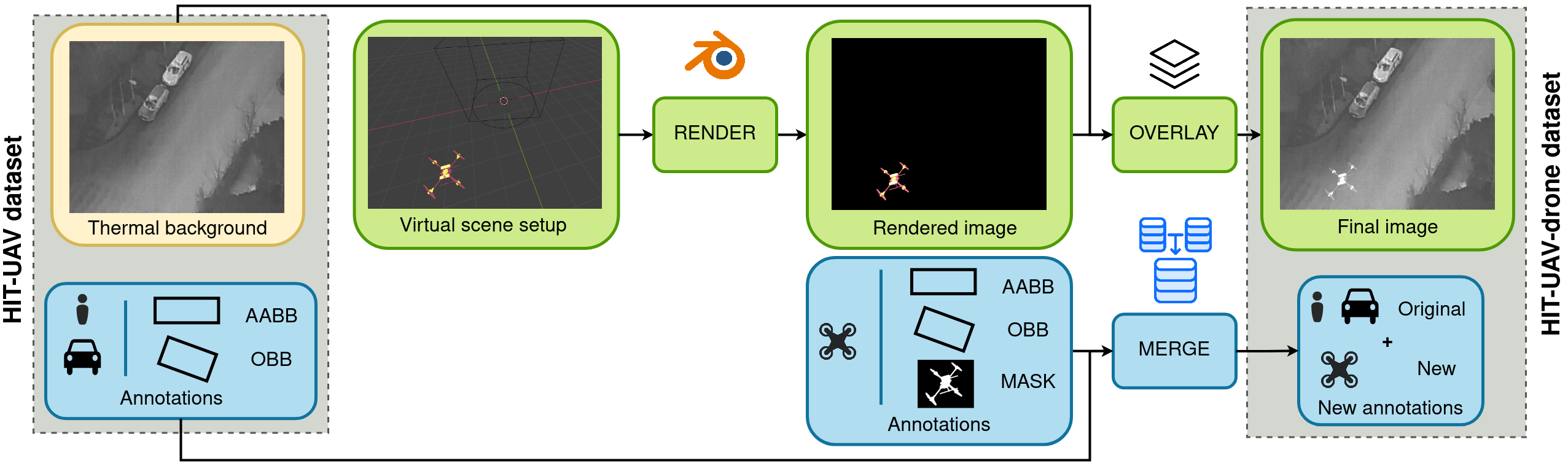}
   \caption{\textbf{Procedural pipeline for generating synthetic thermal images.} Starting from the original HIT-UAV dataset, the thermal background and the corresponding annotations are used as the base. A new virtual scene is procedurally generated, based on configurable parameters, while aligning the camera view with the background's perspective. After rendering new objects, they are overlaid onto the thermal background. Annotations are automatically generated in the form of Axis-Aligned Bounding Boxes (AABB), Oriented Bounding Boxes (OBB), and masks. The final annotations can either be merged with the original annotations or used separately. The Blender-based workflow is highlighted in green, while the annotation-related processes are shown in blue.}
   \label{fig:pipeline}
\end{figure*}

In general, the main goal of this research is to utilize existing thermal aerial datasets and use them for a broader list of tasks than they were originally planned for. Our contributions are threefold: (1) augmenting datasets with new object categories, (2) adding oriented bounding boxes and segmentation masks without extra labeling costs for tasks like oriented detection and segmentation, and (3) increasing dataset diversity through varied object positions, orientations, scales, and additional data for both new and existing classes.


Our pipeline consists of four main steps: 1) thermal background preparation, 2) virtual scene setup, 3) rendering and overlay, and 4) annotation merging, as illustrated in \Cref{fig:pipeline}. The backgrounds are sourced from existing datasets, which include metadata. We determine the camera angles at which the images were captured to maintain accurate perspective. Aerial views offer a unique top-down perspective, which significantly differs from the typical frontal view seen in natural images captured at human eye level. The virtual world creation builds upon our previous work \cite{Barisic2022, Kek2024}, using the same core principles. Once the scene is set up, we render the new objects and overlay them onto the extracted thermal backgrounds. After generating the synthetic images and their corresponding annotations, we merge these with the annotations of the original backgrounds if the categories are intended to be used. During this merging process, we post-process the labels to remove any that are fully overlapped by the new ones. In the following sections, we provide a detailed description of each component.

\subsection{Thermal backgrounds}

There are two options for creating thermal backgrounds that include objects and environments not of primary interest but that provide contextual information: 1) generating a virtual thermal world or 2) using real thermal backgrounds. To minimize the domain gap between synthetic models and real-world data, we chose to use real thermal images. We hypothesize that the data will appear realistic if we can match the aerial viewpoint and render it accurately, which is why we selected Blender \cite{blender} for its high fidelity. For the datasets we used, we extract metadata for each background and incorporate it into the pipeline within the virtual Blender environment.

\subsection{Virtual scene setup}

Dataset diversity enhances both its quality and robustness. To achieve this, we use Blender's Python API to programmatically set up scenes for each image, manipulating objects, textures, cameras, and lighting. We specify the following parameters to configure the pipeline:
\begin{enumerate} 
\item Number of scene configurations per entry: \( N_{\text{config}} \)
\item Distance between the camera and the scene origin: \( D_{\text{min}} \leq d \leq D_{\text{max}} \)
\item Roll angle of the camera: \( \phi_{\text{min}} \leq \phi \leq \phi_{\text{max}} \)
\item Yaw angle of the object: \( \psi_{\text{obj,min}} \leq \psi_{\text{obj}} < \psi_{\text{obj,max}} \)
\item Position of the object in the scene: \( x_{\text{obj,min}} \leq x_{\text{obj}} \leq x_{\text{obj,max}}, \, y_{\text{obj,min}} \leq y_{\text{obj}} \leq y_{\text{obj,max}} \)
\end{enumerate}

%


The object is placed in the scene while the camera follows an animation path based on orientation and distance parameters. Varying these parameters generates diverse scene configurations for our dataset.

\begin{figure}
  \centering
  \subfloat[Deer model]{%
    \includegraphics[width=0.5\columnwidth]{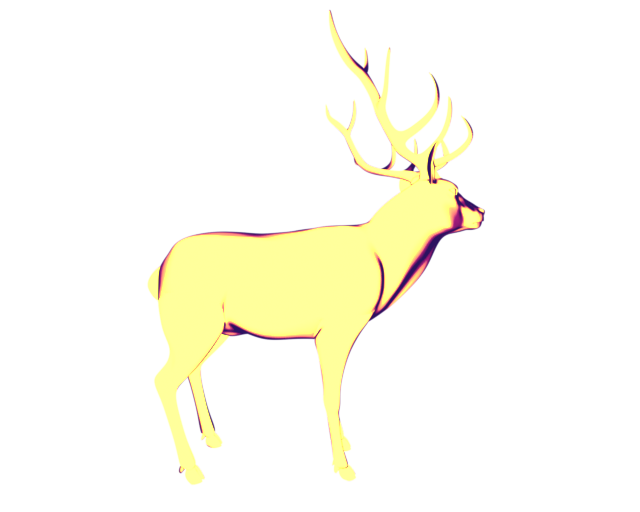}%
    \label{fig:deer}
  }
  \subfloat[Drone model]{%
    \includegraphics[width=0.5\columnwidth]{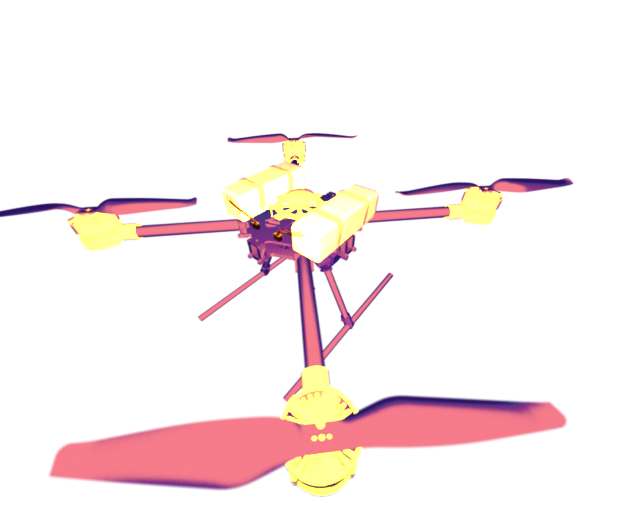}
    \label{fig:crone}
  }
  \caption{3D thermal models used in the procedural pipeline for synthetic thermal data generation.}
  \label{fig:models}
\end{figure}

To simulate thermal emissivity and reflectivity in synthetic objects, we use Blender's Shader Editor to build a node network that emulates realistic thermal effects. Based on real thermal images and general knowledge of the objects, we tailor the node setup to achieve appropriate thermal characteristics. The key node is the InfraredEffect node, which is a custom node that modifies the material's emissivity specifically for thermal imaging. This node ensures that thermal effects are visible from direct angles, accurately reflecting how the object would appear in infrared. It is manually modeled for each object and consists of combination of Emission, Color Ramp, and other material handling nodes. We also use the Fresnel node to adjust the infrared radiation and thermal appearance based on the viewing angle. This aids in creating variations in thermal emissivity as the viewing perspective changes. The thermal 3D models are presented in \Cref{fig:models}. The drone is modeled with multiple surfaces and materials: the batteries and motors are assigned higher temperatures to simulate their heat emission, while the carbon frame and propellers are set to lower temperatures due to their reduced thermal output. In contrast, the deer is modeled more uniformly, representing a consistent temperature distribution across its body.

\subsection{Image render and overlay}

After the virtual scene is set, the image is rendered. To ensure high-quality results, we employ Blender's Cycles engine, which uses photorealistic path-tracing to accurately simulate light behavior. Rendering speed is enhanced by utilizing NVIDIA GPU acceleration via CUDA, leveraging parallel computing power. To integrate the rendered image into the background scenes, we first convert the color thermal images to grayscale to match the format of the input dataset. Using Blender's compositing system, specifically the Alpha Over node, we overlay the rendered layer onto the background image. The script dynamically organizes output images into directories based on parameters such as angle and background settings, streamlining data management.

\begin{figure}[htb]
  \centering
  \subfloat[AABB]{%
    \includegraphics[width=0.31\columnwidth]{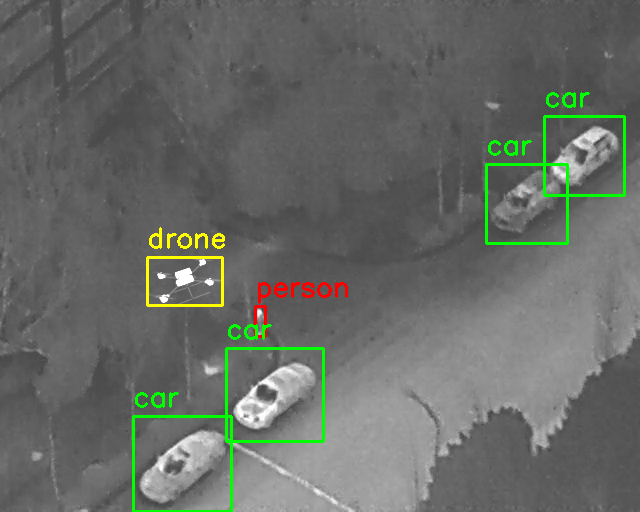}%
    \label{fig:aabb}
  }
  \subfloat[OBB]{%
    \includegraphics[width=0.31\columnwidth]{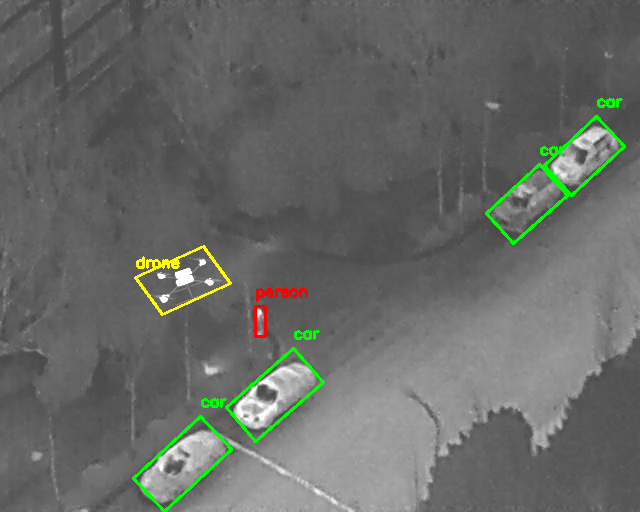}
    \label{fig:obb}
  }
  \subfloat[Mask]{%
    \includegraphics[width=0.31\columnwidth]{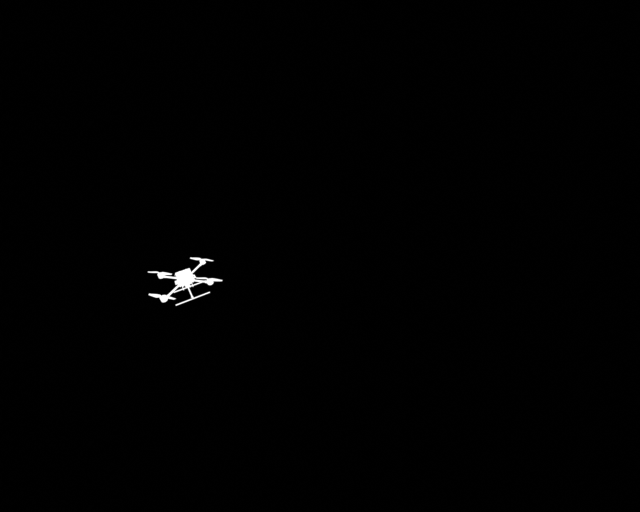}
    \label{fig:seg}
  }
  \caption{Annotation types in the HIT-UAV-drone dataset. We merged annotations for both the original and newly introduced classes, providing AABB and OBB annotations across all classes. Segmentation masks are exclusively available for the newly added class, as they were not included in the original dataset.}
  \label{fig:annotations}
\end{figure}

\subsection{Annotations}

After rendering and compositing the images, two post-processing steps are performed. First, we merge the annotations generated for the synthetic objects with the existing annotations of the original dataset. This involves converting all annotations to a common format (YOLO) and consolidating them into unified annotation files for each image. For the new object class, we provide three types of annotations: Axis-Aligned Bounding Boxes (AABB), Oriented Bounding Boxes (OBB), and pixel-level segmentation masks. In \Cref{fig:annotations} you can see an example from the HIT-UAV-drone dataset. The annotations are generated automatically, using Python scripts in Blender to project 3D objects into the camera frame. The merged annotations are offered in the format defined by the input dataset. Lastly, we verify whether any newly added objects completely obscure original objects. If such cases are detected, the occluded object's labels are removed to ensure that all labeled instances remain visible in the images.


\begin{figure*}[htb]
  \centering
   \includegraphics[width=1.0\linewidth]{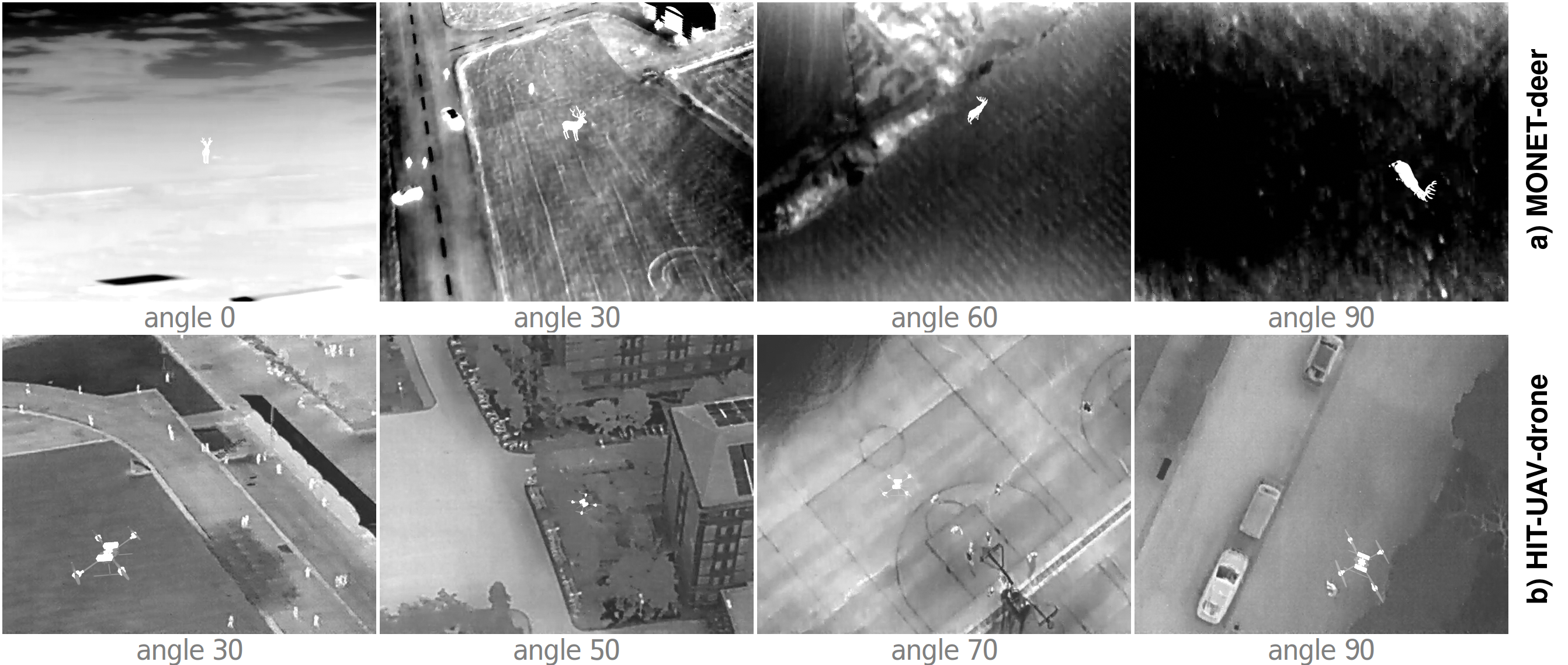}
   \caption{Example data from the synthetically enhanced datasets, showcasing varied viewpoints recreated in a virtual environment using metadata alignment.}
   \label{fig:angles}
\end{figure*}

\section{Experiments}
\label{sec:experiments}


\subsection{Experimental setup}

\textbf{Background datasets.} As previously mentioned, we utilized two thermal background datasets, HIT-UAV and MONET. For the HIT-UAV dataset, we employed the entire set of images: $2,029$ for training, $579$ for testing, and $290$ for validation. Each image includes metadata specifying the camera angle, which varies between $30^\circ$ and $90^\circ$. During label analysis of the MONET dataset, we found that $21.64\%$ of the training set metadata is missing, primarily from the subsets \textit{runway\_03}, \textit{runway\_02}, \textit{runway\_01}, and \textit{dirtroad\_01}. To maintain consistency and avoid potential issues in model training and evaluation, we excluded these subsets. In the MONET dataset, the camera is mounted on a gimbal, with the pitch angle ranging from $-40.0^\circ$ to $90.0^\circ$. The few instances of negative gimbal pitch likely stem from sudden UAV movements, which the gimbal system could not fully compensate for. Both the test and validation sets contain complete metadata and were used without further modifications. 



\textbf{Virtual scene parameters.} Before initiating the procedural pipeline in Blender, we configure the scene settings and define parameter ranges to ensure a diverse dataset. The image resolution is set to $640 \times 512$, using the Cycles rendering engine with $640$ samples per image. The distance between the camera and the scene origin, $d$, is randomized within a range of $1$ to $10$ meters, while object's $x_{\text{obj}}$ and $y_{\text{obj}}$ coordinates are varied between $-3$ and $3$ meters, resulting in a variable distance between camera and object. Due to the randomized placement of the camera and object, the object may occasionally fall outside the camera’s field of view. To simulate the dynamic conditions of a moving drone, a random roll $\phi$ between $-10^\circ$ and $10^\circ$ is introduced to the camera. The object’s yaw orientation $\psi_{\text{obj}}$ is randomized across the full range, from $0^\circ$ to $360^\circ$. These parameters are randomized for each image, ensuring that every scene is unique. For the HIT-UAV dataset, we generate two virtual scenes ($N_{\text{config}}=2$) for each background, followed by rendering and compositing. In contrast, for the larger MONET dataset, we limit the setup to one scene per background to manage the substantial dataset size.

\subsection{New enhanced datasets}

    
    
    



We introduce two new datasets that enhance existing datasets: MONET-deer and HIT-UAV-drone. The names indicate the new classes added to each dataset. Details of these datasets are presented in \Cref{tab:new_datasets}. In MONET-deer, we show that a new class, important for animal monitoring and protection in rural areas, namely deer, can be synthetically generated and overlayed on a real background, removing the need for a new collection of the data. We automatically and with precision provide AABB, OBB and segmentation masks for annotation of the newly added class. We keep the dataset size, as its is already substantial. Example images from the new dataset, taken from variety of angles and illustrating the full range of aerial perspectives covered in this pipeline, are shown in \Cref{fig:angles}. For the HIT-UAV-drone dataset, we incorporate the "drone" class, which is crucial for urban mobility and safety. Similar to the MONET-deer dataset, the new class is synthetically generated and integrated with existing thermal data. To further enhance the dataset, we double its size by generating two virtual scenes for each background. This demonstrates the effectiveness of our technique in expanding the dataset’s volume without the need for new data collection, a method that can also be applied to existing categories. For both datasets we remove the "don't care" class as YOLO does not support ignoring specific regions within an image, which is the intended purpose of that class.

\begin{table}[h!]
\centering
\caption{Comparison of original and enhanced datasets}
\setlength{\tabcolsep}{3pt} 
\begin{tabular}{lccc}
\toprule
\textbf{Dataset}       & \textbf{\# Images} & \textbf{\# Classes} & \textbf{Annotation Type} \\ 
\midrule
MONET           & 32248              & 2         & AABB                     \\ 
MONET-deer      & 32248              & 3         & AABB, OBB, Masks         \\ 
\addlinespace
HIT-UAV         & 2898               & 4         & AABB, OBB              \\ 
HIT-UAV-drone   & 5796               & 5         & AABB, OBB, Masks         \\
\bottomrule
\end{tabular}
\label{tab:new_datasets}
\end{table}


\subsection{Object detection}

We utilize newly generated datasets by training a common object detection model YOLOv8 on them. Specifically, we used the "s" model variant, with a learning rate set to $0.002$, a batch size of $64$, and an image resolution of $640$, while keeping all other parameters at their default values. The training process was configured for a maximum of $1000$ epochs, with early stopping applied using a patience of $100$ epochs. In the table \Cref{tab:yolov8_hituavdrone} we presents results of training on the HIT-UAV-drone dataset and evaluating on the test split of the dataset. The overall performance demonstrates strong detection capabilities, with a mean Average Precision (mAP) of $0.898$ at $0.5$ Intersection over Union (IoU). The newly added class is the top-performing category with mAP@0.5 of $0.995$, indicating that the detector is effectively trained on the proposed synthetic enhancements of the thermal dataset. Classes such as "person" and "bicycle" are more challenging to detect due to their smaller size, while the "Other Vehicle" class lacks sufficient data for the model to learn the robust representation.

\begin{table}[htbp]
\centering
\caption{Test results on HIT-UAV-drone dataset}
\setlength{\tabcolsep}{2pt} 
\begin{tabular}{lcccccc}
\toprule
 Class &  Images &  Instances &     P &     R &  mAP50 &  mAP50-95 \\
\midrule
   All           & 1158  & 10721 & 0.916 & 0.873 & 0.898 & 0.673 \\
Person           & 710   & 5222  & 0.905 & 0.884 & 0.923 & 0.512 \\
Car              & 534   & 2678  & 0.943 & 0.943 & 0.965 & 0.750 \\
Bicycle          & 172   & 1592  & 0.911 & 0.839 & 0.907 & 0.598 \\
Other Vehicle    & 42    & 68    & 0.824 & 0.706 & 0.699 & 0.541 \\
Drone            & 1158  & 1161  & 0.997 & 0.994 & 0.995 & 0.964 \\
\bottomrule
\end{tabular}
\label{tab:yolov8_hituavdrone}
\end{table}

On a substantially larger dataset MONET-deer, we further demonstrate the effectiveness of our approach by introducing the new class "deer." The results on the test part of the dataset are presented in \Cref{tab:yolov8_monetdeer}. The YOLOv8 model achieved an overall mAP@50 of $0.821$, with the "deer" class attaining the highest mAP@0.5 of $0.930$. The "person" class remains challenging to detect due to the small area of corresponding bounding boxes.

\begin{table}[htbp]
\centering
\caption{Test results on MONET-deer dataset}
\setlength{\tabcolsep}{3.5pt} 
\begin{tabular}{lcccccc}
\toprule
 Class &  Images &  Instances &     P &     R &  mAP50 &  mAP50-95 \\
\midrule
   all   &    7224   &      36983   & 0.892 & 0.797 &  0.821 &     0.569 \\
person   &    5999   &      20940   & 0.842 & 0.629 &  0.661 &     0.354 \\
   car   &    6254   &       8819   & 0.859 & 0.867 &  0.873 &     0.469 \\
  deer   &    7224   &       7224   & 0.976 & 0.897 &  0.930 &     0.884 \\
\bottomrule
\end{tabular}
\label{tab:yolov8_monetdeer}
\end{table}

\subsection{Ablation studies}

\textbf{From visible-light to thermal images.} We trained the same object detection model, YOLOv8s, on the DUT-Anti-UAV dataset, which consists exclusively of aerial visible-light images annotated for a single class: drone. We then compare the performance of the detector trained on visible-light images with that of the detector trained on our enhanced synthetic thermal dataset, focusing on the drone class. The comparison results of training on visible light and thermal images are presented in \Cref{tab:visible_thermal}. The detector trained on our HIT-UAV-drone dataset achieves a significantly higher mAP@0.5 of $0.995$ compared to $0.429$ for the DUT-Anti-UAV-trained model, highlighting the significant domain gap between visible and thermal imaging. This disparity is anticipated due to inherent differences in object features from two different modalities. Also, visible light images contain more texture information can be used for detection, and thermal images have lower resolution. Additionally, the DUT-Anti-UAV dataset predominantly captures front and ground-level perspectives, which likely further diminishes detection performance when applied to our datasets, recorded from higher altitudes with overhead viewpoints.

\begin{table}[h]
\centering
\caption{Comparison of visible light and thermal training data on the thermal test data with "drone" class}
\setlength{\tabcolsep}{8pt} 
\begin{tabular}{lccccccc}
\hline
Train data  & P & R & mAP50 & mAP50-95 \\
\hline
DUT-Anti-UAV    & 0.582 & 0.47  & 0.429 & 0.254 \\
HIT-UAV-drone   & 0.997 & 0.994 & 0.995 & 0.964 \\
\hline
\end{tabular}
\label{tab:visible_thermal}
\end{table}

\textbf{Comparison with the original dataset.} We train object detector on the original dataset HIT-UAV to compare performance with the new dataset on joint classes. Comparing results from \Cref{tab:yolov8_hituavdrone} and \Cref{tab:yolov8_hit_uav_original} we conclude that the addition of new class does not affect the performance on the other classes. The differences in metrics are small, and both positive and negative depending on the metric. It’s important to note that the extended dataset includes an additional class, potential occlusions between the new class and existing data, and twice as many images, so some variation is expected.

\begin{table}[htbp]
\centering
\caption{Test results on HIT-UAV dataset}
\setlength{\tabcolsep}{2pt} 
\begin{tabular}{lcccccc}
\toprule
Class        & Images & Instances & P & R & mAP50 & mAP50-95 \\
\midrule
All          & 579    & 4780      & 0.895 & 0.858 & 0.884 & 0.606 \\
Person       & 355    & 2611      & 0.898 & 0.881 & 0.925 & 0.520 \\
Car          & 267    & 1339      & 0.948 & 0.950 & 0.972 & 0.761 \\
Bicycle      & 86     & 796       & 0.902 & 0.867 & 0.920 & 0.600 \\
Other Vehicle& 21     & 34        & 0.833 & 0.735 & 0.717 & 0.545 \\
\bottomrule
\end{tabular}
\label{tab:yolov8_hit_uav_original}
\end{table}

\textbf{Aerial perspective significance.} To demonstrate importance of aerial perspective in the datasets used for training deep learning models for aerial applications, we generate two more versions of the HIT-UAV-drone dataset: with fixed pitch angle of the camera at $0^\circ$ and another with randomly varied angles. Again, we train object detector YOLOv8 on new modifications of the training data. The comparison of detectors trained on three different versions of training data, and tested on test part of the HIT-UAV-drone with metadata-aligned angle are presented in \Cref{tab:angle_comparison}. When the camera angle is fixed at eye level ($0^\circ$), the new class is rendered from a frontal view, which does not accurately represent the diverse angles typical in aerial perspectives. This mismatch results in poor performance, which is reflected in a mAP@0.5 of $0.464$. In contrast, training with random pitch angles that encompass the full range of aerial perspectives results in comparable performance to our proposed method. Nevertheless, the highest performance is achieved using our proposed method, which precisely aligns the virtual scene objects and camera angles with real-world values used while capturing background thermal images.

\begin{table}[ht]
\centering
\caption{Comparison of fixed, random and metadata aligned pitch angle in procedural pipeline during generation of HIT-UAV-drone dataset}
\setlength{\tabcolsep}{6pt} 
\begin{tabular}{lcccc}
\hline
\textbf{Angle} & \textbf{P} & \textbf{R} & \textbf{mAP@50} & \textbf{mAP@50-95} \\ \hline
Fixed at $0^\circ$  & 0.518 & 0.551 & 0.464 & 0.133 \\
Random [$0^\circ$,$90^\circ$] & 0.985 & 0.972 & 0.981 & 0.910 \\
Metadata aligned & 0.997 & 0.994 & 0.995 & 0.964 \\ \hline
\end{tabular}
\label{tab:angle_comparison}
\end{table}

\section{Conclusions, Limitations and Future Work}
\label{sec:conclusion}

This study shows that adding new classes to existing thermal datasets broadens the applications of thermal aerial imagery without affecting existing class performance. We highlight a domain gap when transferring deep learning models from visible light to thermal imagery, emphasizing the need for dedicated thermal data. We also demonstrate that aligning the aerial perspective improves thermal image analysis. For future work, we plan to incorporate more classes, including technical entities such as power lines and tower, and various animal species. This is one of the downsides of our pipeline, as we require 3D models of objects with modeled thermal shading. We also aim to explore multimodal integration, leveraging existing thermal-visual pairs.


\addtolength{\textheight}{-12cm}   








\bibliographystyle{ieeetr}
\bibliography{bibliography/bib}

\end{document}